\documentclass[11pt,a4paper]{article}
\usepackage[hyperref]{acl2021}
\usepackage{times}
\usepackage{latexsym}
\usepackage{multirow}
\usepackage{graphicx}
\usepackage{booktabs}
\usepackage{bm}

\usepackage{microtype}
\usepackage{fdsymbol}
\usepackage{colortbl}
\definecolor{Blue2}{RGB}{235, 245, 250}

\aclfinalcopy 
\def\aclpaperid{378}

\title{GLGE: A New General Language Generation Evaluation Benchmark}

\author{Dayiheng Liu$^\spadesuit{\,}$\thanks{\hspace{2mm}Work is done during internship at Microsoft Research Asia.} \:\:\: \textbf{Yu Yan}$^{\dag}{\,}^{\heartsuit}$ \:\:\: \textbf{Yeyun Gong}$^{\dag}{\,}^{\clubsuit}$ \:\:\:  \textbf{Weizhen Qi}$^{\clubsuit}$ \:\:\: \textbf{Hang Zhang}$^{\spadesuit}$
\\ 
\textbf{Jian Jiao}$^{\heartsuit}$ \:\:\: \textbf{Weizhu Chen}$^{\ddag}$ \:\:\:\: \textbf{Jie Fu}$^\diamondsuit$ \:\:\:\: \textbf{Linjun Shou}$^{\vardiamondsuit}$ \:\:\:\:  \textbf{Ming Gong}$^{\vardiamondsuit}$
\\ 
\textbf{Pengcheng Wang}$^{\heartsuit}$ \:\:\:\:  
\textbf{Jiusheng Chen}$^{\heartsuit}$  \:\:\:\:  \textbf{Daxin Jiang}$^{\vardiamondsuit}$  \:\:\: \textbf{Jiancheng Lv}$^{\spadesuit}$ \\
\textbf{Ruofei Zhang}$^{\heartsuit}$ \:\:\: \textbf{Winnie Wu}$^{\heartsuit}$ \:\:\: \textbf{Ming Zhou}$^{\clubsuit}$ \:\:\: \textbf{Nan Duan}$^{\clubsuit}$ \\
$^\spadesuit$College of Computer Science, Sichuan University    \:\:\:\:  $^\clubsuit$Microsoft Research Asia  \:\:\:\: $^\diamondsuit$ Mila \\  
$^\ddag$Microsoft Azure AI
$^\vardiamondsuit$Microsoft Search Technology Center Asia $^\heartsuit$Microsoft \\ \texttt{glge@microsoft.com} \:\:\:\: $^\dag$Equal contribution
}

\date{}

\begin{document}
\maketitle
\begin{abstract}
Multi-task benchmarks such as GLUE and SuperGLUE have driven great progress of pretraining and transfer learning in Natural Language Processing (NLP). These benchmarks mostly focus on a range of Natural Language Understanding (NLU) tasks, without considering the Natural Language Generation (NLG) models. 
In this paper, we present the \textbf{G}eneral \textbf{L}anguage \textbf{G}eneration \textbf{E}valuation (\textbf{GLGE}), a new multi-task benchmark for evaluating the generalization capabilities of NLG models across eight language generation tasks. For each task, we continue to design three subtasks in terms of task difficulty (\textbf{GLGE-Easy}, \textbf{GLGE-Medium}, and \textbf{GLGE-Hard}). This introduces 24 subtasks to comprehensively compare model performance. To encourage research on pretraining and transfer learning on NLG models, we make GLGE publicly available and build a leaderboard with strong baselines including MASS, BART, and ProphetNet\footnote{The source code and dataset are publicly available at \url{https://github.com/microsoft/glge}.}.
\end{abstract}

\section{Introduction}
Pretrained language models, such as BERT~\cite{devlin2018bert} and other advanced pretrained models~\cite{raffel2019exploring,yang2019xlnet,liu2019roberta,alberti2019bert,brown2020language,clark2020electra} have made great progress in a host of Natural Language Understanding (NLU) tasks. Meanwhile, the development of general evaluation benchmarks has also helped drive the progress of these models. These benchmarks usually use an overall score to evaluate the performance of models across a wide range of NLU tasks. In addition to GLUE~\cite{wang2018glue} and SuperGLUE~\cite{wang2019superglue} which are general language understanding evaluation benchmarks for English, several general language understanding evaluation benchmarks for other languages are proposed, such as CLUE~\cite{xu2020clue} for Chinese, FLUE~\cite{le2019flaubert} for French, and IndoNLU~\cite{wilie2020indonlu} for Indonesian. Furthermore, the multilingual multi-task benchmarks such as XTREME~\cite{hu2020xtreme} and XGLUE~\cite{liang2020xglue} are proposed for cross-lingual evaluation. 

In addition to NLU tasks, an increasing number of pretrained language models designed for Natural Language Generation (NLG) tasks have recently been proposed, such as MASS~\cite{song2019mass}, BERT-share~\cite{rothe2020leveraging}, BART~\cite{lewis2019bart}, ProphetNet~\cite{yan2020prophetnet}, and ERINE-GEN~\cite{xiao2020ernie}. However, the generalization capabilities of the language generation of these models are usually evaluated with different tasks, datasets, and metrics, which cannot provide a coherent and comprehensive evaluation. Although there are several general evaluation benchmarks as we mentioned above, none of them are particularly designed for general language generation evaluation. 

To fill the gap of the NLG evaluation benchmark, we introduce the \textbf{G}eneral \textbf{L}anguage \textbf{G}eneration \textbf{E}valuation (\textbf{GLGE}) benchmark, a new multi-task benchmark for evaluating the generalization capabilities of NLG in English language. It contains eight English language generation tasks, covering text summarization, question generation, generative question answering, and dialogue. We select six pre-existing popular datasets and introduce two new datasets selected from real-world scenarios. Moreover, in order to provide more diversified difficulty challenges, we employ two simple but effective strategies to build three NLG evaluation benchmarks (called \textbf{GLGE-Easy}, \textbf{GLGE-Medium}, and \textbf{GLGE-Hard}) in terms of task difficulty. 

To better understand the challenges posed by GLGE, we conduct experiments with existing widely used non-pretrained models (e.g., vanilla LSTM Seq2Seq~\cite{bahdanau2014neural}, vanilla Transformer~\cite{vaswani2017attention}), and pretrained models (e.g., MASS~\cite{song2019mass}, BART~\cite{lewis2019bart}, and ProphetNet~\cite{yan2020prophetnet}). We further analyze the n-gram diversity of the output samples. The experimental results show that there is a large performance gap between the pretrained models and the non-pretrained models. However, on the GLGE-hard task, the performance of the pretrained models still has great room for improvement.

In summary, the contributions of this work are five-fold: (1) a new multi-task NLG evaluation benchmark consisting of eight distinct datasets across four kinds of typical NLG tasks, (2) three NLG evaluation benchmarks of different difficulty levels, (3) standardized evaluation metrics and scripts for model evaluation and comparison, (4) open-sourced baselines and a public leaderboard\footnote{\url{https://microsoft.github.io/glge/}.} for the benchmark, (5) a thorough comparative study on existing widely used non-pretrained models and pretrained models with a detailed analysis of the results.

\section{GLGE Benchmark}

\subsection{Design Principles}
For the GLGE benchmark, we design and select the NLG tasks based on the following principles:

\subsubsection{Task Diversity}
The tasks in GLGE focus on evaluating the generalization capabilities of a NLG model, varying the task, the length of the input text, the length of the output text, the type of generated text, and the size of the dataset.

\subsubsection{Task Difficulty}\label{sec:level}
The tasks in GLGE should be challenging but solvable, which can encourage researchers to design better NLG models. Furthermore, we aim to provide benchmarks of different difficulty levels like GLUE~\cite{wang2018glue} and SuperGLUE~\cite{wang2019superglue}, which allows researchers to comprehensively evaluate the models. Researchers can also select the benchmark with moderate difficulty according to the size of the model and the scale of the used pretraining corpus for comparison.

\subsubsection{Ease of Evaluation}
The tasks in GLGE should be easily evaluated automatically. For some unconditional, open-ended, and weak conditional language generation tasks (e.g., answer-agnostic question generation, single-turn chit-chat response generation, and story generation), reasonable generation results are diverse. Due to the limited number of references in the automatic evaluation of text generation tasks, it is more difficult for automatic evaluation of those tasks. Therefore, instead of selecting unconditional and weak conditional language generation tasks, we tend to select language generation tasks with stronger conditions (e.g., answer-aware question generation), which makes the automatic evaluation more convincing.

\subsubsection{Task Popularity}
Most tasks in GLGE should use widely-used NLG datasets, which have been implicitly agreed upon by the NLG community as challenging and meaningful. Since GLGE is mainly designed for the generalization capabilities evaluation of the English NLG pretrained model, the choice of task also refers to several related works of NLG pre-training model, such as MASS~\cite{song2019mass}, BART~\cite{lewis2019bart}, ProphetNet~\cite{yan2020prophetnet}, and ERNIE-GEN~\cite{xiao2020ernie}.

\begin{table*}[t]
\small
\centering
\begin{tabular}{lrrrrrccc}
\toprule
\textbf{Corpus} & $|$\textbf{Train}$|$ & $|$\textbf{Dev}$|$ & $|$\textbf{Test}$|$ & $|$\textbf{Src.}$|$ & $|$\textbf{Tgt.}$|$ & \textbf{Input} & \textbf{Output} & \textbf{Metric} \\ 
\midrule
\multicolumn{9}{c}{\bf Abstractive Text Summarization}\\
\midrule
CNN/DailyMail & 287,113  & 13,368 & 11,490 & 822.3 & 57.9 & article & summary & R-1/R-2/R-L   \\
Gigaword & 3,803,957  & 189,651  & 1,951 & 33.7 & 8.7 & passage & headline & R-1/R-2/R-L   \\
XSUM & 204,017  & 11,327 & 11,333 & 358.5 & 21.1 & article & summary & R-1/R-2/R-L  \\
MSNews & 136,082  & 7,496  & 7,562 & 310.7 & 9.7 & article & headline & R-1/R-2/R-L   \\ \midrule
\multicolumn{9}{c}{\bf Answer-aware Question Generation}\\
\midrule
SQuAD 1.1 & 75,722 & 10,570 & 11,877 & 149.4 & 11.5 & answer/passage & question &  R-L/B-4/MTR \\
MSQG & 198,058 & 11,008 & 11,022  & 45.9 & 5.9 & highlight/passage & question &  R-L/B-4/MTR\\
\midrule
\multicolumn{9}{c}{\bf Conversational Question Answering}\\
\midrule
CoQA & 108,647 & 3,935 & 4,048  & 354.4 & 2.6 & history/passage & answer &  F1-Score  \\ \midrule
\multicolumn{9}{c}{\bf Personalizing Dialogue}\\
\midrule
PersonaChat & 122,499 & 14,602 & 14,056 & 120.8 & 11.8 & persona/history & response &  B-1/B-2/D-1/D-2 \\ 
\bottomrule
\end{tabular}
\caption{GLGE task descriptions and statistics. $|$Train$|$: the number of examples in train set. $|$Src.$|$: the average number of words in source inputs. R-L: ROUGE-L. B-4: BLUE-4. MTR: METEOR. D-2: Distinct-2.}
\label{tab:glge}
\end{table*}

Based on the above principles, we invite 10 NLG experts\footnote{Each expert has experience in publishing multiple papers at top conferences in the NLG field.} to discuss and vote on existing widely-used NLG datasets. Note that since the GLGE is designed for evaluating the generalization capabilities of NLG in English language, we do not include the cross-lingual NLG tasks, such as machine translation and cross-lingual text summarization~\cite{liang2020xglue}. Finally, we select 6 existing popular NLG datasets. Besides, we also introduce two new datasets selected from real-world scenarios for the GLGE benchmark, which makes GLGE have more practical values. Unlike the existing datasets, the test sets of these new two datasets are hidden, which further ensures the fairness of the evaluation results. The input sequence and output sequence of the selected tasks are all well-defined. We pre-process them and provide the input and output sequence pairs directly, which benefits researchers to focus on model improvements.

\subsection{Tasks and Datasets} \label{sec:task}
GLGE contains eight English NLG tasks, covering text summarization, question generation, generative question answering, and dialogue. Descriptions and statistics of these tasks are shown in Table~\ref{tab:glge}, with concrete 
examples shown in Appendix.

\subsubsection{Abstractive Text Summarization}
As a typical NLG task, abstractive text summarization aims to generate a short and fluent summary of a long text document. GLGE contains four abstractive text summarization tasks. As discussed in~\citet{bhandari2020re}, we use ROUGE-1, ROUGE-2, and ROUGE-L~\cite{lin2004rouge} as the metrics for these tasks. 

\noindent\textbf{CNN/DailyMail}~\citep{hermann2015cnndm} dataset contains 220K articles from the Daily Mail newspapers and 93K articles from the CNN. Each article contains a bullet point summary. GLGE uses the non-anonymized variant \citet{see2017get}. After the pre-processing, there are 311,971 $\langle$article, summary$\rangle$ pairs, where the source input is the article, and the target output is the summary which consists of multiple sentences.

\noindent\textbf{Gigaword} \citep{rush2015neural} contains 4M examples extracted from the news articles of the Gigaword corpus \citep{graff2003gigaword}. After the pre-processing, there are 3,995,559 $\langle$passage, summary$\rangle$ data pairs, where the source input is the first sentence of the article, and the target output is the headline that usually contains a single sentence. 

\noindent\textbf{XSum} \citep{narayan2018don} consists of 227K online articles from the British Broadcasting Corporation (BBC), which contains professionally written single-sentence summaries. After the pre-processing, there are 226,677 $\langle$article, summary$\rangle$ data pairs, where the source input is the news article, and the target output is a single-sentence summary. 

\noindent\textbf{MSNews} \textbf{M}icro\textbf{S}oft \textbf{News} headline generation (\textbf{MSNews}) is a new News headline generation dataset we collected for GLGE. We random select 151K online news articles from 2012-01-01 to 2020-09-01 from a real-world news search engine. Each article contains a professionally written single-sentence headline. After the pre-processing, there are 151,140 $\langle$article, headline$\rangle$ data pairs, where the source input is the news article, and the target output is a news headline.

\subsubsection{Answer-aware Question Generation}
The question generation task is another typical NLG task, which aims to generate a question based on a given text passage or document. Compared with answer-agnostic question generation tasks that can generate lots of reasonable questions, answer-aware question generation~\cite{zhou2017neural} is asked to generate a question asks towards the given answer span based on a given text passage or document. In order to facilitate automatic evaluation, GLGE selects two answer-aware question generation tasks:

\noindent\textbf{SQuAD 1.1} \citep{rajpurkar2016squad} dataset contains over 100K crowd-worker created questions with the corresponding answer spans in 536 Wikipedia articles. Since the original hidden test set of the SQuAD 1.1 is hidden, we follow~\cite{du2017learning,zhao2018paragraph} to re-split the dataset with the examples from the original training set and development set. After the pre-processing, there are 98,169 $\langle$answer, passage, question$\rangle$ data triples, in which the source input is a Wikipedia passage along with an answer span, and the target output is a question. ROUGE-L, BLEU-4~\cite{papineni2002bleu}, and METEOR~\cite{banerjee2005meteor} are used as the metrics.

\noindent\textbf{MSQG} \textbf{M}icro\textbf{S}oft \textbf{Q}uestion \textbf{G}eneration (\textbf{MSQG}) is another dataset we collected, which is a new challenge dataset, the questions in this dataset are freely edited by daily users. For MSQG, we collect 220K passages from a real world search engine. Each passage contains a highlight span and a related query, we regard the queries as questions in this dataset. After the pre-processing, there are 220,088 $\langle$highlight span, passage, question$\rangle$ data triples, where the source input is a news passage along with highlight span, and the target output is a user question. ROUGE-L, BLEU-4, and METEOR are used as the metrics.  

\subsubsection{Conversational Question Answering}
Conversational question answering is a classic and popular generative question answering task. Compared with the extractive question answering, such as SQuAD~\cite{rajpurkar2016squad}, conversational question answering requires the model to answer the question based on a running conversation history and the given passage.

\noindent\textbf{CoQA}~\cite{reddy2019coqa} dataset contains 127K questions with answers, obtained from 8K conversations about text passages from seven diverse domains. After the pre-processing, there are 116,630 $\langle$conversation history, passage, question, answer$\rangle$ data 4-tuples, where the source input is a sequence of conversation history along with a given question and a given passage, and the target output is a free-form answer text. F1-Score~\cite{rajpurkar2016squad} is used as the metric.

\subsubsection{Personalized Dialogue}
Conversational AI is an important topic in NLG. Compared with text summarization, the responses of single-turn conversations are diverse and might lack of specification, and thus it is hard to use the single ground-truth for automatic evaluation. We select the personalizing dialogue task, which is a challenging multi-turn conversation task. In addition to the conversation history, this task gives the profile information as an additional condition to facilitate specific response generation.

\noindent\textbf{PersonaChat}~\cite{zhang2018personalizing} dataset consists of about 160K utterances, which requires the model to generate responses according to given multi-turn conversations and persona profile. After pre-processing, there are 151,157 $\langle$persona profile description text, conversation history, response$\rangle$ data triples, where the source input is a sequence of conversation history along with several sentences of persona profile description information, and the target output is a response. BLEU-1, BLEU-2, Distinct-1, and Distinct-2~\cite{li2016diversity} are used as the evaluation metrics.

\subsection{Overall Score}
\label{sec:over_score}
Similar to GLUE~\cite{wang2018glue} and SuperGLUE~\cite{wang2019superglue}, we seek to give an overall system performance over all GLGE tasks by aggregating the scores of all tasks. We follow GLUE to adopt a simple approach that weighs each task equally. For the tasks with multiple metrics, we firstly average those metrics to get a task score. Besides, because the values of the original Distinct-1 (D-1) and Distinct-2 (D-2)~\cite{li2016diversity} scores which are used as the metrics for dialogue task are usually quite small (less than 0.01), we re-scale them by 100.0 so that these score values are in the same order of magnitude as other scores.

\subsection{Challenges of Three Difficulty Levels}\label{sec:difficult}
As discussed in \S~\ref{sec:level}, GLGE provides three levels of difficulty for each task, called \textbf{GLGE-Easy}, \textbf{GLGE-Medium}, and \textbf{GLGE-Hard}. The original 8 task datasets as described in \S~\ref{sec:task} constitute the GLGE-Easy. Based on GLGE-Easy, we employ two strategies to further increase the task difficulty.

\noindent\textbf{Low-resource}. We increase the difficulty of GLGE by simulating low-resource scenarios. For each task, we keep the test and development sets of GLGE-Easy and randomly reduce the scale of the training data to 50\% of the original train set. The dataset of 8 tasks under this setting is regarded as GLGE-Medium.

\noindent\textbf{Low-frequency}. In order to further evaluate the generalization capability of the NLG model, we increase the difficulty of GLGE by reducing the word overlap rate between the output of the training set and the output of the test set. The motivation is that a good NLG model should be able to generate a fluent target output based on the input information, even if the target output may contain some low-frequency words. For the test set and development sets of GLGE-Hard, we still use those in GLGE-Easy. For the training set of GLGE-Hard, we first count the frequency of each token in the target sentence of the test set. Then we remove the stop words of the target sentence of each training sample in GLGE-Easy, ranking them by calculating their word frequency score of the test set. This can be formulated as
\begin{equation}
\textrm{Score}(y) = \frac{\sum_{w_y \in y}{\textrm{TF}(w_y)}}{|y|},
\end{equation}
where $y$ is a target sentence without stop words, $w_y$ is a token in $y$, $\textrm{TF}(w_y)$ denotes the word frequency of the token $w_y$ in the target sentences of the whole test set, and $|y|$ denotes the token length of $y$. Instead of reducing the training data scale randomly as in GLGE-Medium, we select the top 25\% training data with minimum word frequency score of the test set from the original training set as the training set of each dataset. The dataset of 8 tasks under this setting is regarded as GLGE-Hard.

\begin{table*}[!ht]
\resizebox{1 \textwidth}{!}{%
\begin{tabular}{l c cc cc ccc c}
\toprule
\multirow{2}{*}{\textbf{Models}} & \multirow{2}{*}{\textbf{Avg.}} & \multicolumn{4}{c}{Text Summarization} & \multicolumn{2}{c}{Question Generation} & \multicolumn{1}{c}{QA} & \multicolumn{1}{c}{Dialogue} \\\
 & & \textbf{CNN/DM} & \textbf{Gigaword} & \textbf{XSUM} & \textbf{MSNews} & \textbf{SQuAD 1.1} & \textbf{MSQG} & \textbf{CoQA} & \textbf{PesonaChat} \\
\midrule
\textbf{Metrics} & &\multicolumn{4}{c}{R-1/R-2/R-L} & \multicolumn{2}{c}{R-L/B-4/MTR} & F1 & B-1/B-2/D-1/D-2\\ \midrule
\multicolumn{10}{c}{\textbf{GLGE-Easy}} \\ \midrule
LSTM & \cellcolor{Blue2}{20.0} & 37.3/15.7/34.4 & 34.2/16.0/31.8 & 25.1/6.9/19.9 & 30.0/14.6/27.7 & 27.2/3.8/8.9 & 25.3/3.5/14.1 & 15.1 & 42.2/35.9/0.2/0.7 \\
Transformer & \cellcolor{Blue2}{21.9} & 39.5/16.7/36.7 & 37.1/18.4/34.5 & 30.5/10.4/24.2 & 33.0/15.4/30.0 & 30.7/4.8/10.9 & 29.3/5.1/16.6 & 15.7 & 38.3/33.6/0.2/0.7 \\
MASS\textsubscript{base} & \cellcolor{Blue2}{33.6} & 42.1/19.5/39.0 & 38.7/19.7/35.9 & 39.7/17.2/31.9 & 39.4/21.0/36.1 & 49.4/20.1/24.4 & 38.9/10.2/23.3 & 65.4 & 41.0/35.7/1.4/6.9 \\
ProphetNet\textsubscript{base} & \cellcolor{Blue2}{33.8} & 42.5/19.7/39.5 & 38.9/19.9/36.0 & 39.8/17.1/32.0 & 40.6/21.6/37.0 & 48.0/19.5/23.9 & 37.1/9.3/22.7 & 65.3 & 46.0/38.4/1.3/7.3 \\
MASS\textsubscript{middle} & \cellcolor{Blue2}{34.3} & 42.9/19.8/39.8 & 38.9/20.2/36.2 & 39.1/16.5/31.4 & 40.4/21.5/36.8 & 49.9/21.3/25.2 & 38.9/9.5/23.5 & 67.6 & 46.0/38.2/1.2/6.2 \\ 
BART\textsubscript{large} & \cellcolor{Blue2}{35.8} & 44.1/21.2/40.9 & 38.1/18.4/34.9 & 45.1/22.2/37.2 & 43.8/24.0/39.2 & 50.3/22.0/26.4 & 38.8/9.2/24.3 & 68.6 & 49.9/40.0/1.3/8.0 \\
ProphetNet\textsubscript{large} & \cellcolor{Blue2}{36.5} & 44.2/21.1/41.3 & 39.5/20.4/36.6 & 44.4/21.3/36.4 & 44.1/24.4/40.2 & 51.5/22.5/26.0 & 38.3/9.6/23.3 & 73.0 & 46.7/39.0/1.3/7.5 \\ \midrule
\multicolumn{10}{c}{\textbf{GLGE-Medium}} \\ \midrule
LSTM & \cellcolor{Blue2}{18.1} & 35.3/14.1/32.8 & 33.3/15.2/31.1 & 21.5/4.6/17.1 & 27.0/12.1/24.9 & 26.6/3.5/8.2 & 18.6/1.7/9.5 & 12.9 & 41.3/35.3/0.1/0.5\\
Transformer & \cellcolor{Blue2}{19.5} & 35.0/11.0/32.4 & 36.7/18.1/34.1 & 27.5/8.3/21.8 & 26.8/9.7/24.3 & 28.3/4.1/9.8 & 27.0/4.2/15.0 & 14.2 & 37.7/29.6/0.2/0.7\\
MASS\textsubscript{base} & \cellcolor{Blue2}{33.0} & 41.2/18.8/38.2 & 37.9/19.1/35.2 & 37.4/14.9/29.8 & 38.9/20.5/35.6 & 48.9/20.0/24.3 & 38.2/9.5/22.8 & 65.0 & 42.8/36.7/1.3/6.2 \\ 
ProphetNet\textsubscript{base} & \cellcolor{Blue2}{32.6} & 41.6/19.2/38.7 & 38.6/19.6/35.7 & 37.8/15.3/30.4 & 39.0/20.4/35.7 & 46.4/17.9/22.5 & 37.0/8.7/22.3 & 62.5 & 45.4/37.7/1.4/7.3 \\
MASS\textsubscript{middle} & \cellcolor{Blue2}{33.6} & 41.5/19.0/38.5 & 38.3/19.1/35.4 & 38.4/15.8/30.7 & 39.6/20.9/36.0 & 49.3/20.4/24.4 & 38.3/9.9/22.7 & 67.2 & 44.0/37.3/1.3/6.1 \\ 
BART\textsubscript{large} & \cellcolor{Blue2}{35.3} & 42.8/20.1/39.1 & 38.0/18.3/34.7 & 43.1/19.5/34.1 & 43.4/23.6/38.9 & 49.7/21.6/25.9 & 38.4/9.5/24.0 & 69.4 & 50.4/39.1/1.2/7.4 \\
ProphetNet\textsubscript{large} & \cellcolor{Blue2}{35.5} & 43.1/20.3/40.1 & 39.1/19.8/36.1 & 41.8/18.7/33.8 & 43.3/23.5/39.4 & 50.4/21.9/25.8 & 39.3/10.0/23.7 & 72.3 & 42.0/36.4/1.4/7.8 \\ \midrule
\multicolumn{10}{c}{\textbf{GLGE-Hard}} \\ \midrule
LSTM & \cellcolor{Blue2}{12.6} & 26.2/6.8/24.2 & 26.3/9.2/24.6 & 17.8/2.4/14.3 & 8.2/0.9/7.6 & 27.3/1.0/6.7 & 12.5/0.4/5.0 & 10.3 & 36.8/28.7/0.1/0.4\\
Transformer & \cellcolor{Blue2}{14.4} & 28.3/6.2/25.8 & 28.6/10.8/26.5 & 23.0/5.3/18.3 & 18.0/3.5/16.2 & 25.9/1.1/7.0 & 17.0/1.3/8.2 & 9.9 & 30.0/29.7/0.1/0.2\\
MASS\textsubscript{base} & \cellcolor{Blue2}{28.2} & 40.4/18.0/37.3 & 32.2/13.6/29.5 & 33.7/11.6/26.7 & 35.4/17.0/32.4 & 42.8/13.4/19.0 & 34.1/7.5/18.6 & 50.2 & 40.1/34.9/1.6/7.8 \\ 
ProphetNet\textsubscript{base} & \cellcolor{Blue2}{28.0} & 40.9/18.4/37.7 & 32.0/13.5/29.5 & 34.2/11.6/26.8 & 35.2/17.0/32.1 & 41.6/13.4/18.9 & 32.3/7.2/18.0 & 48.5 & 41.6/35.5/1.6/8.3 \\
MASS\textsubscript{middle} & \cellcolor{Blue2}{29.1} & 41.1/18.5/38.0 & 32.2/13.5/29.9 & 34.9/12.5/27.6 & 36.6/18.0/33.4 & 45.1/16.0/21.3 & 34.3/8.0/19.0 & 51.2 & 41.4/35.4/1.5/7.6 \\ 
BART\textsubscript{large} & \cellcolor{Blue2}{31.0} & 41.7/19.1/37.9 & 33.0/13.6/30.0 & 39.7/16.1/30.9 & 40.8/20.8/36.4 & 45.9/18.1/23.7 & 35.1/8.5/20.7 & 53.5 & 48.3/37.3/1.3/7.2 \\
ProphetNet\textsubscript{large} & \cellcolor{Blue2}{30.5} & 41.2/18.7/38.0 & 32.4/13.7/29.9 & 39.4/16.1/31.6 & 40.3/20.5/36.4 & 46.4/17.0/22.1 & 34.0/8.2/19.0 & 54.1 & 40.5/35.2/1.8/9.2 \\ 
\bottomrule
\end{tabular}%
}
\caption{
Overall results of baselines on all GLGE tasks. We use the \colorbox{Blue2}{color} to highlight the overall score. R-1: ROUGE-1. R-2: ROUGE-2. R-L: ROUGE-L. B-4: BLUE-4. MTR: METEOR. D-1: Distinct-1. D-2: Distinct-2. Note that as discussed in \S~\ref{sec:over_score}, the values of Distinct-1 and Distinct-2 are multiplied by 100.
}\label{tab:main-results}
\end{table*}

\section{Experiments}
\subsection{Baselines}~\label{sec:baselines}
For the baselines, we first evaluate two widely-used non-pretrained models: vanilla LSTM based Seq2Seq~\cite{bahdanau2014neural} and vanilla Transformer~\cite{vaswani2017attention}. Besides, we evaluate several widely used pretrained NLG models, including MASS~\cite{song2019mass}, BART~\cite{lewis2019bart}, and ProphetNet~\cite{yan2020prophetnet}. To further evaluate the performance of the pretrained NLG models of different model sizes and the different scales of the pretraining corpus, we compare the MASS\textsubscript{base}, ProphetNet\textsubscript{base}, MASS\textsubscript{middle}, BART\textsubscript{large}, and ProphetNet\textsubscript{large} on GLGE.

\subsection{Implementation Details}~\label{sec:settings}

\noindent\textbf{Vanilla LSTM}~\cite{bahdanau2014neural}. The hyper-parameters and implementation of LSTM-Seq2Seq are based on the \textit{LSTM} register model of Fairseq\footnote{\url{https://github.com/pytorch/fairseq/blob/master/fairseq/models/lstm.py}.}, where the word embedding dimension, the hidden size, the number of the encoder layer, and the number of the decoder layer are 512, 512, 1, and 1, respectively. For each task in GLGE, we use Adam~\cite{kingma2014adam} with an initial learning rate of between 0.0001 and 0.0003, and train the LSTM-Seq2Seq for a maximum of 100 epochs.

\noindent\textbf{Vanilla Transformer}~\cite{vaswani2017attention}. The hyper-parameters and implementation of Transformer are based on the \textit{transformer\_vaswani\_wmt\_en\_de\_big} register model of Fairseq\footnote{\url{https://github.com/pytorch/fairseq/blob/master/fairseq/models/transformer.py}.}, which contains a 6-layer encoder and a 6-layer decoder with 1024 embedding/hidden size and 4096 feed-forward filter size. For each task in GLGE, we use Adam with the initial learning rate of between 0.0003 and 0.001, and train the Transformer for a maximum of 20 epochs.

\noindent\textbf{MASS}\textsubscript{base}~\cite{song2019mass}. The hyper-parameters and implementation of MASS are based on their source code\footnote{\url{https://github.com/microsoft/MASS}}. MASS\textsubscript{base} contains a 6-layer encoder and a 6-layer decoder with 768 embedding/hidden size and 3072 feed-forward filter size. The MASS\textsubscript{base} is pretrained on BookCorpus~\cite{zhu2015aligning} and English Wikipedia (16GB in total). For each task in GLGE, we fine-tune MASS\textsubscript{base} with the same hyper-parameters used in their source code\footnote{\url{https://github.com/microsoft/MASS/tree/master/MASS-summarization}} for a maximum of 25 epochs. 

\noindent\textbf{MASS}\textsubscript{middle}~\cite{song2019mass} which contains a 6-layer encoder and a 6-layer decoder with 1024 embedding/hidden size and 4096 feed-forward filter size. The MASS\textsubscript{middle} is also pretrained on BookCorpus and English Wikipedia (16GB in total). For each task in GLGE, we use the same hyper-parameters as used in MASS\textsubscript{base}.

\noindent\textbf{ProphetNet}\textsubscript{base}~\cite{yan2020prophetnet}. The hyper-parameters and implementation of ProphetNet are based on their source code\footnote{\url{https://github.com/microsoft/ProphetNet}}. ProphetNet\textsubscript{base} contains a 6-layer encoder and a 6-layer decoder with 768 embedding/hidden size and 3072 feed-forward filter size. Similar to MASS, the ProphetNet\textsubscript{base} is pretrained on BookCorpus and English Wikipedia (16GB in total) with 125K steps. For each task in GLGE, we fine-tune ProphetNet\textsubscript{base} with the same hyper-parameters used in their source code for a maximum of 10 epochs.

\noindent\textbf{ProphetNet}\textsubscript{large}~\cite{yan2020prophetnet}. It contains a 12-layer encoder and 12-layer decoder with 1024 embedding/hidden size and 4096 feed-forward filter size. The ProphetNet\textsubscript{large} is pretrained on the 160GB English language corpora of news, books, stories, and web text for 14 epochs. For each task in GLGE, we fine-tune ProphetNet\textsubscript{large} with the same hyper-parameters used in their source code for a maximum of 10 epochs.

\noindent\textbf{BART}\textsubscript{large}~\cite{lewis2019bart}. The hyper-parameters and implementation of BART\textsubscript{large} are based on the source code\footnote{\url{https://github.com/pytorch/fairseq/tree/master/examples/bart}}. BART\textsubscript{large} contains a 12-layer encoder and 12-layer decoder with 1024 embedding/hidden size and 4096 feed-forward filter size. The pretraining of BART\textsubscript{large} uses the same pretraining data as \citet{liu2019roberta}, consisting of 160GB of news, books, stories, and web text. For each task in GLGE, we fine-tune BART\textsubscript{large} with the same hyper-parameters used in their source code\footnote{\url{https://github.com/pytorch/fairseq/blob/master/examples/bart/README.summarization.md}} for a maximum of 20000 iterations.

Except BART, all the baselines adopt BERT-uncased tokenizer. We fine-tune all baselines on each individual task with $4 \times 16$GB NVIDIA V100 GPUs. We evaluate the best model checkpoint based on the loss on the development set. During inference, we use beam search~\cite{och2004alignment} with beam size 4 or 5 and remove the duplicated trigrams in beam search~\cite{fan2017controllable} to obtain the generated results.

\subsection{Results and Analysis}
\noindent\textbf{Overall results}.
The main results are presented in Table~\ref{tab:main-results}. From the overall scores (highlighted in color), we can observe the fairly consistent gains moving from LSTM to Transformer, and then to pretrained base models and pretrained large models, such as ProphetNet\textsubscript{base} and ProphetNet\textsubscript{large}. The performance gap between the pretrained model and non-pretrained model is obvious. The difference in terms of overall score is about absolute 15\% on the three levels GLGE benchmarks (GLGE-Easy, GLGE-Medium, and GLGE-Hard). As expected, the pretrained large models (ProphetNet\textsubscript{large} and BART\textsubscript{large}) achieve the best overall scores. 

\begin{table*}
\resizebox{1 \textwidth}{!}{
\begin{tabular}{lccccccccc}
\toprule
\multirow{2}{*}{\textbf{Models}} & \multirow{2}{*}{\textbf{Avg.}} & \multicolumn{4}{c}{Text Summarization} & \multicolumn{2}{c}{Question Generation} & \multicolumn{1}{c}{QA} & \multicolumn{1}{c}{Dialogue} \\
 & & \textbf{CNN/DM} & \textbf{Gigaword} & \textbf{XSUM} & \textbf{MSNews} & \textbf{SQuAD 1.1} & \textbf{MSQG} & \textbf{CoQA} & \textbf{PesonaChat} \\\midrule
\textbf{Metrics} & &\multicolumn{4}{c}{R-1/R-2/R-L} & \multicolumn{2}{c}{R-L/B-4/MTR} & F1 & B-1/B-2/D-1/D-2\\\midrule
\multicolumn{10}{c}{\textbf{GLGE-Medium + Low-frequency Strategy}} \\\midrule
LSTM & \cellcolor{Blue2}{16.4} & 35.4/13.9/32.7 & 31.3/13.2/29.3 & 20.8/4.0/16.5 & 25.3/10.4/23.3 & 24.8/1.2/6.8 & 15.3/1.1/7.1 & 10.5 & 34.8/31.6/0.2/0.6\\
Transformer & \cellcolor{Blue2}{18.1} & 36.8/13.7/34.0 & 33.2/14.6/30.9 & 25.9/7.3/20.7 & 25.2/7.6/22.7 & 26.4/1.9/7.8 & 22.8/2.7/11.9 & 10.8 & 39.6/34.3/0.1/0.2\\
MASS\textsubscript{base} & \cellcolor{Blue2}{29.9} & 41.3/18.8/38.3 & 34.4/15.3/31.7 & 36.1/13.6/28.5 & 36.4/17.9/33.1 & 45.3/16.0/21.4 & 34.6/8.5/19.9 & 54.6 & 40.3/34.6/1.5/7.5 \\ 
ProphetNet\textsubscript{base} & \cellcolor{Blue2}{29.9} & 41.7/19.1/38.7 & 35.1/16.1/32.4 & 36.5/14.0/28.9 & 37.4/18.6/34.0 & 43.7/15.1/20.5 & 33.8/7.6/20.0 & 53.1 & 42.3/35.9/1.4/7.6 \\
MASS\textsubscript{middle} & \cellcolor{Blue2}{30.8} & 41.5/18.9/38.4 & 35.1/16.0/32.5 & 37.0/14.4/29.4 & 37.9/19.1/34.6 & 46.3/17.0/22.1 & 35.5/8.9/20.5 & 56.6 & 41.0/35.5/1.6/8.0 \\ 
BART\textsubscript{large} & \cellcolor{Blue2}{32.0} & 42.2/19.6/38.4 & 35.7/16.1/32.6 & 41.5/17.9/32.3 & 41.5/21.4/37.0 & 46.5/18.1/24.5 & 35.9/8.4/21.7 & 54.6 & 49.1/38.1/1.3/8.4 \\
ProphetNet\textsubscript{large} & \cellcolor{Blue2}{32.3} & 42.7/20.0/39.5 & 35.3/16.2/32.6 & 40.6/17.4/32.6 & 40.9/21.4/37.1 & 47.8/19.2/23.9 & 36.4/9.4/20.9 & 57.6 & 44.4/36.9/1.4/8.3 \\\bottomrule
\end{tabular}
}
\caption{
Overall results of baselines across the tasks of GLGE-Medium + Low-frequency strategy. We use the \colorbox{Blue2}{color} to highlight the overall score. R-1: ROUGE-1. R-2: ROUGE-2. R-L: ROUGE-L. B-4: BLUE-4. MTR: METEOR. D-1: Distinct-1. D-2: Distinct-2. 
}
\label{tab:ab-results}
\end{table*}

From the results of each model on the three levels GLGE benchmarks, we can see that each model has a significant drop in performance from GLGE-Easy to GLGE-Medium and GLGE-Hard benchmarks. For both the non-pretrained model and pretrained models, there is a nearly 2\% drop in terms of overall score from GLGE-Easy to GLGE-Medium, and about 4\%-8\% drop from GLGE-Easy to GLGE-Hard. These results illustrate the diversified difficulty of GLGE. We recommend that researchers choose a GLGE benchmark with moderate difficulty based on the model size of the pretrained model and the scale size of the pretrained corpus. At the same time, researchers can also perform more comprehensive evaluation of model performance on GLGE benchmarks at all difficulty levels.  

\noindent\textbf{Low-frequency strategy analysis}.
To further verify the effectiveness of the low-frequency strategy as described in \S~\ref{sec:difficult}. We build the GLGE-Medium + low-frequency benchmark and evaluate the models on it. Both GLGE-Medium and GLGE-Medium + low-frequency retain 50\% of the training samples in the GLGE-Easy training set. The only difference between them is that GLGE-Medium uses random sampling to retain 50\% training samples, while GLGE-Medium + low-frequency uses the low-frequency strategy as used in GLGE-Hard to select 50\% training samples. We use the same baselines with the same settings in \S~\ref{sec:settings} to compare the model performance on GLGE-Medium and GLGE-Medium + low-frequency. The results are shown in Table~\ref{tab:ab-results}. We can see that after the introduction of the low-frequency strategy, the performance of the models has dropped significantly. These results demonstrate that the low-frequency strategy can effectively improve the difficulty of the benchmark.

\begin{figure*}[!ht]
    \centering
    \includegraphics[width = 6.2in]{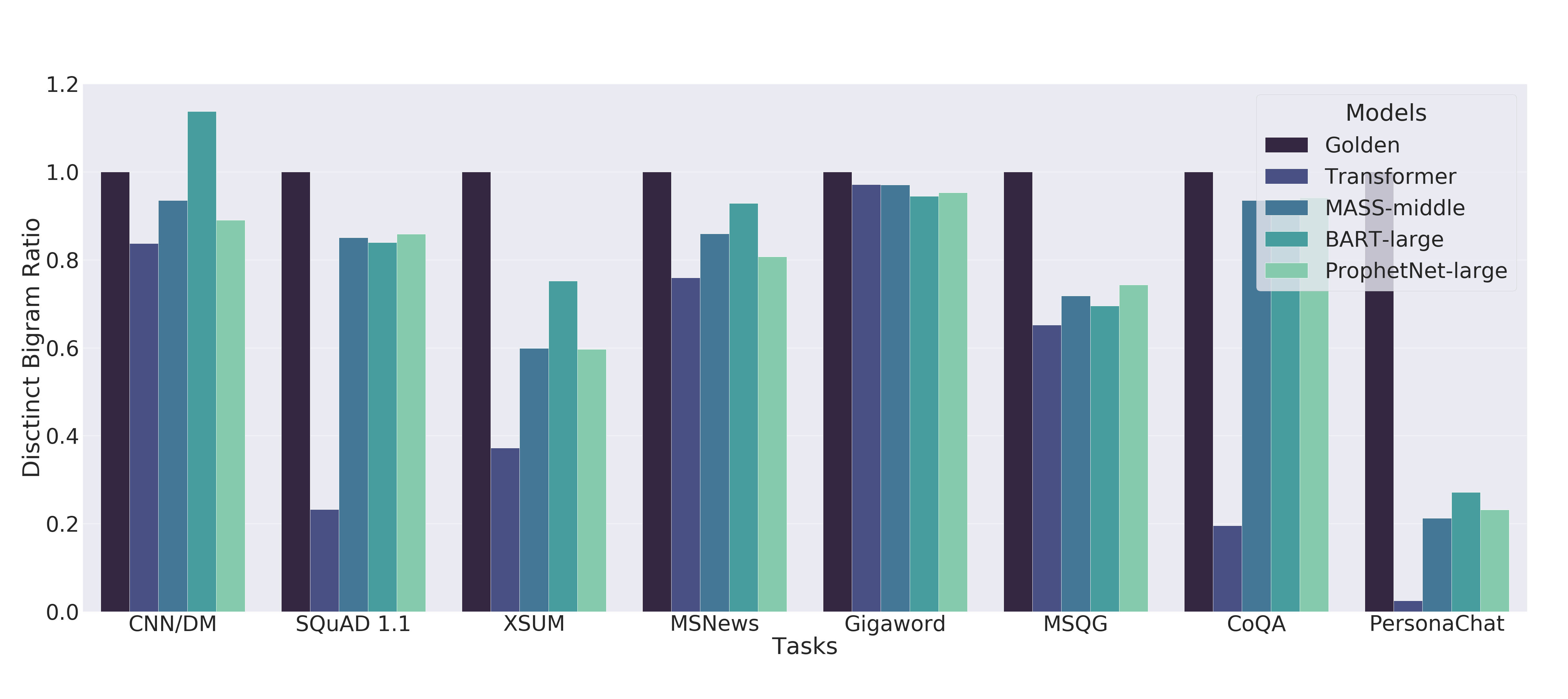}
    \caption{Distinct bigram ratios of the generated samples to the golden references on GLGE-Easy.}
    \label{fig:ngram_easy}
\end{figure*}

\noindent\textbf{Output diversity analysis}.
We further compare the output diversity of each model on all the tasks of GLGE-Easy. We report the mean of the Distinct bigram (Distinct-2)~\cite{li2016diversity} ratios of the generated samples to the golden references. Note that if the bigram diversity of the generated samples is close to that of the real samples, the distinct bigram ratio is close to 1. The results are shown in Figure~\ref{fig:ngram_easy}. In general, output bigram diversity of the pretrained model is higher than non-pretrained models. For the tasks of CNN/DailyMail (CNN/DM), Gigaword, and MSNews, the bigram diversity of the generated samples are close to that of the real samples. However, for the tasks of XSUM, SQuAD 1.1, MSQG, CoQA, and PersonaChat, the bigram diversity of the non-pretrained model is significantly lower than that of the pretrained model. For these tasks, the non-pretrained model tends to generate universal responses~\cite{li2016diversity} or outputs. Moreover, there is still a huge gap between the bigram diversity of pretrained models and real samples (golden) on the task of XSUM, MSQG, and PersonaChat. 
Obviously, there exists great room for future improvement of the pretrained models in terms of output diversity.

\section{Related Works}
\paragraph{Benchmarks} Recently, the development of general natural language understanding (NLU) evaluation benchmarks has helped drive the progress of pretraining and transfer learning in NLP.
\citet{conneau2018senteval} propose a toolkit, SentEval, for evaluating the quality of universal sentence representations.
DecaNLP~\cite{mccann2018natural} casts ten diversified NLP tasks as a general question-answering format for evaluation.
\citet{wang2018glue} propose a widely-used multi-task benchmark, GLUE, for NLU in the English language. There are nine NLU tasks in GLUE, including two single-sentence tasks, three similarity and paraphrase tasks, and four natural language inference tasks.
After that, SuperGLUE~\cite{wang2019superglue} is proposed as a harder counterpart of GLUE. Besides sentence- and sentence-pair classification tasks used in GLUE, SuperGLUE extends the task formats by introducing coreference resolution and question answering tasks.
More recently, A new NLU benchmark called DialoGLUE~\cite{mehri2020dialoglue} for task-oriented dialogue is proposed. It consists of seven task-oriented dialogue datasets covering four kinds of NLU tasks: intent prediction, slot tagging, semantic parsing, and dialogue state tracking.

In addition to the English NLU evaluation benchmark, there has been an increasing amount of new benchmarks in other languages. 
For example, CLUE~\cite{xu2020clue} is a Chinese NLU benchmark that consists of eight diverse Chinese NLU tasks, including single-sentence, sentence-pair, and machine reading comprehension tasks.
FLUE~\cite{le2019flaubert} is proposed for the French language, which is a French NLU benchmark that includes several NLU tasks, such as text classification, paraphrasing, language inference, parsing, POS tagging, and word sense disambiguation.
IndoNLU~\cite{wilie2020indonlu} is a new benchmark for evaluating Indonesian language understanding, which introduces twelve tasks, ranging from single sentence classification to pair-sentences sequence labeling.
Furthermore, the multilingual multi-task benchmarks are proposed for cross-lingual evaluating. 
\citet{hu2020xtreme} introduce XTREME benchmark which is a multi-task benchmark for evaluating the cross-lingual generalization capabilities of multilingual representations across forty languages and nine tasks. Almost at the same time, XGLUE~\cite{liang2020xglue} is proposed which is a new multilingual multi-task benchmark for cross-lingual pretraining, understanding, and generation. There are eleven cross-lingual tasks including nine NLU tasks and two NLG tasks in XGLUE, each task provides labeled data in multiple languages. 

The above benchmarks mostly focus on NLU which provides a range of language understanding tasks. However, to our best knowledge, there is no benchmark designed specifically for general NLG evaluation. To fill this gap, we introduce GLGE, a new multi-task benchmark for evaluating the generalization capabilities of NLG across eight language generation tasks. 

\paragraph{Pretrained NLG Models} 
In recent years, pretrained language models~\cite{devlin2018bert,raffel2019exploring,yang2019xlnet,liu2019roberta,alberti2019bert,brown2020language,clark2020electra} have achieved state-of-the-art results in several NLU benchmarks.
Besides, more and more pretraining based models which are designed for NLG tasks are proposed.
\citet{rothe2020leveraging} adopt the Transformer-based sequence-to-sequence (seq2seq) model and leverage the checkpoints of the pretrained NLU models for sequence generation tasks.
MASS~\cite{song2019mass} pretrains the seq2seq model by dropping a continuous token span to corrupt the text and learns to reconstruct it. 
\citet{raffel2019exploring} investigate several model structures and pretraining tasks, and further propose a unified
text-to-text transformer called T5.
Similarly, BART~\cite{lewis2019bart} adopts the encoder-decoder structure and is pretrained with randomly sentence order reconstruction and text in-filling tasks.

More recently, \citet{yan2020prophetnet} propose, ProphetNet, which introduces the future n-gram prediction mechanism for language generation. ENRINE-GEN~\cite{xiao2020ernie} introduces the infilling generation mechanism, noise-aware generation, and span-by-span generation task for NLG model pretraining. 
In addition to general NLG tasks, some task-specific pretrained NLG models are proposed. 
For dialogue and conversation, a dialogue generative pretrained transformer called DialoGPT~\cite{zhang2019dialogpt} is proposed for conversational response generation, which is pretrained on a large-scale conversation-like exchange corpus.
Furthermore, PLATO~\cite{bao2019plato} is a dialogue generation pretraining framework for chit-chat, knowledge grounded dialogues, and conversational question answering. PLATO introduces discrete latent variables to tackle the one-to-many mapping problem in response generation.
For text summarization, \citet{zhang2019pegasus} propose PEGASUS, which design the pretraining objectives called gap sentence generation tailored for abstractive text summarization.

\section{Conclusion}
To facilitate the development, evaluation, and comparison of new NLG models, we introduce GLGE, a multi-task evaluation benchmark for NLG with three difficulty levels. To the best of our knowledge, GLGE is the first comprehensive NLG evaluation benchmark. 
We evaluate several baselines on GLGE and analyze their results. 
The GLGE benchmark is hosted publicly and we invite the research community to submit to the leaderboard.

In future work, we will try to introduce other automatic evaluation metrics, such as BERTscore~\cite{zhang2019bertscore} and BLEURT~\cite{sellam2020bleurt}. Besides, we will compare the correlation between these metrics and human judgment.

\section*{Acknowledgments}
This work was supported in part by the National Key R\&D Program of China under Grant 2020YFB1406702, in part by NFSC under Grant 61625204, 61836006, and the Science and Technology Major Project of Sichuan province under Grant 2020YFG0478. 
We thank STC-A NLP, Bing Answers, Bing Ads, Bing Relevance and Microsoft News for providing the datasets.

\bibliographystyle{acl_natbib}
\bibliography{example_output}

\appendix

\begin{table*}
\centering \footnotesize
\begin{tabular}{p{0.005\textwidth}p{0.93\textwidth}}
 \toprule
 \parbox[t]{1mm}{\multirow{2}{*}{\rotatebox[origin=c]{90}{{\textbf{CNN/DailyMail}}}}} &
\textbf{Article:} \textit{The only thing crazier than a guy in snowbound Massachusetts boxing up the powdery white stuff and offering it for sale online? People are actually buying it. For \$ 89, self-styled entrepreneur Kyle Waring will ship you 6 pounds of Boston-area snow in an insulated Styrofoam box -- enough for 10 to 15 snowballs, he says. [...], a coastal suburb north of Boston. He joked about shipping the stuff to friends and family in warmer states, and an idea was born [...]} \\ & \textbf{Target:} \underline{A man in suburban Boston is selling snow online to customers in warmer states. For \$ 89, he will ship 6 pounds} \\ & \underline{of snow in an insulated Styrofoam box.} \\
\midrule
\parbox[t]{1mm}{\multirow{3}{*}{\rotatebox[origin=c]{90}{{\textbf{Gigaword}}}}} &
\textbf{Passage:} \textit{U.S. business leaders lashed out Wednesday at legislation that would penalize companies for employing illegal immigrants.} \\& \\ & \textbf{Target:} \underline{U.S. business attacks tough immigration law.} \\ 
\midrule
\parbox[t]{1mm}{\multirow{2}{*}{\rotatebox[origin=c]{90}{{\textbf{XSUM}}}}} &
\textbf{Article:} \textit{Burberry reported pre-tax profits of £166m for the year to March. [...], Sales rose 7\% to £1.28bn, with the company recording double-digit sales growth in Europe and Asia Pacific. Adjusted profit rose 23\% to £215m, taking into account one-off items and a favourable exchange rate. Stores in London in particular benefited from favourable currency movements and increased tourism. [...], Burberry shares were up 7.6\% at 659 pence in afternoon trading.} \\ & \textbf{Target:} \underline{Luxury fashion designer Burberry has returned to profit after opening new stores and spending more on online} \\
&\underline{marketing.}\\
\midrule
\parbox[t]{1mm}{\multirow{2}{*}{\rotatebox[origin=c]{90}{{\textbf{MSNews}}}}} &
\textbf{Article:} \textit{Los Angeles : Actor Chadwick Boseman, who played Black icons Jackie Robinson and James Brown before finding fame as the regal Black Panther in the Marvel cinematic universe, died Friday of cancer, his representative said. He was 43. Boseman died at his home in the Los Angeles area with his wife and family by his side, his publicist Nicki Fioravante told The Associated Press. [...]} \\ & \textbf{Target:} \underline{Black Panther actor Chadwick Boseman dies of cancer at 43.} \\
\midrule
\parbox[t]{1mm}{\multirow{2}{*}{\rotatebox[origin=c]{90}{{\textbf{SQuAD 1.1}}}}} & 
\textbf{Passage:} \textit{Super Bowl 50 was an American football game to determine the champion of the National Football League (NFL) for the 2015 season. The American Football Conference (AFC) champion Denver Broncos defeated the National Football Conference (NFC) champion Carolina Panthers 24–10 to earn their third Super Bowl title. The game was played on February 7, 2016, at Levi's Stadium in the San Francisco Bay Area at Santa Clara, California. As this was the 50th Super Bowl, [...].} \\ & \textbf{Answer:} \textit{Santa Clara, California} \\ & \textbf{Target:} \underline{Where did Super Bowl 50 take place?} \\
\midrule
\parbox[t]{1mm}{\multirow{2}{*}{\rotatebox[origin=c]{90}{{\textbf{MSQG}}}}} &
\textbf{Passage:} \textit{On March 28, 1830, Congress passed the Indian Removal Act, beginning the forced relocation of thousands of Native Americans in what became known as the Trail of Tears. Not all members of Congress supported the Indian Removal Act.} \\ & \textbf{Highlight Span:} \textit{Indian Removal Act}\\ & \textbf{Target:} \underline{What act was passed to relocate the native Americans?}\\
\midrule
\parbox[t]{1mm}{\multirow{2}{*}{\rotatebox[origin=c]{90}{{\textbf{CoQA}}}}} &
\textbf{Passage:} \textit{John was in the third grade, and nine years old. Every day he had to walk home from school. There were some kids in his class who were mean to him, and during the winter they would throw snowballs at him. [...], John thought it was a good deal, and ended up being much better at math.} \\ & \textbf{Conversation History:} \textit{Q1: Who is in third grade? A1: John. Q2: How old is he? A2: Nine.} \\ & \textbf{Question:} \textit{What did kids do to him?} \\ & \textbf{Target:} \underline{Throw snowballs at him.} \\
\midrule
\parbox[t]{1mm}{\multirow{1}{*}{\rotatebox[origin=c]{90}{{\textbf{PersonaChat}}}}} & 
\textbf{Persona Profile description text:} \textit{My wife left me and took my children. I don't believe in god. I'm overweight and unhappy. I work at a nursing home. I spend most of my time on Facebook when I'm not working.} \\ &
\textbf{Conversation History:} \textit{Q1: I got a big house with 7 rooms. R1: Nice, majority of my time I am on Facebook one time. Q2: I saw a man fly to the moon.}
 \\ & \textbf{Target:} \underline{Like on TV? I'm obese and unhappy.} \\
\bottomrule
\end{tabular}
\caption{Development-set examples from the tasks in GLGE. \textbf{Bold} text denotes part of the example format for each task. Text in \textit{italics} is part of the source input text. \underline{Underlined} text is the target output text. \textit{[...]} denotes the omitted texts.}
\label{tab:data_example}
\end{table*}

\end{document}